\documentclass[10pt,twocolumn,letterpaper]{article}

\usepackage{iccv}
\usepackage{times}
\usepackage{epsfig}
\usepackage{graphicx}
\usepackage{amsmath}
\usepackage{amssymb}
\usepackage{float}
\usepackage{dblfloatfix}

\usepackage[breaklinks=true,bookmarks=false]{hyperref}

\iccvfinalcopy

\def\iccvPaperID{12696}

\ificcvfinal\fi

\begin{document}

%%%%%%%%% TITLE
\title{EyeBAG: Accurate Control of Eye Blink and Gaze Based on Data Augmentation Leveraging Style Mixing}

\author{Bryan S Kim\\
KAIST\\
{\tt\small bryanswkim@kaist.ac.kr}
\and
Jeong Young Jeong\\
Innerverz\\
{\tt\small pensee0.0a@innerverz.com}
\and
Wonjong Ryu\\
Innerverz\\
{\tt\small 1zong2@innerverz.com}
}

\makeatletter
\g@addto@macro\@maketitle{
  \begin{figure}[H]
  \begin{center}
  \includegraphics[width=\textwidth]{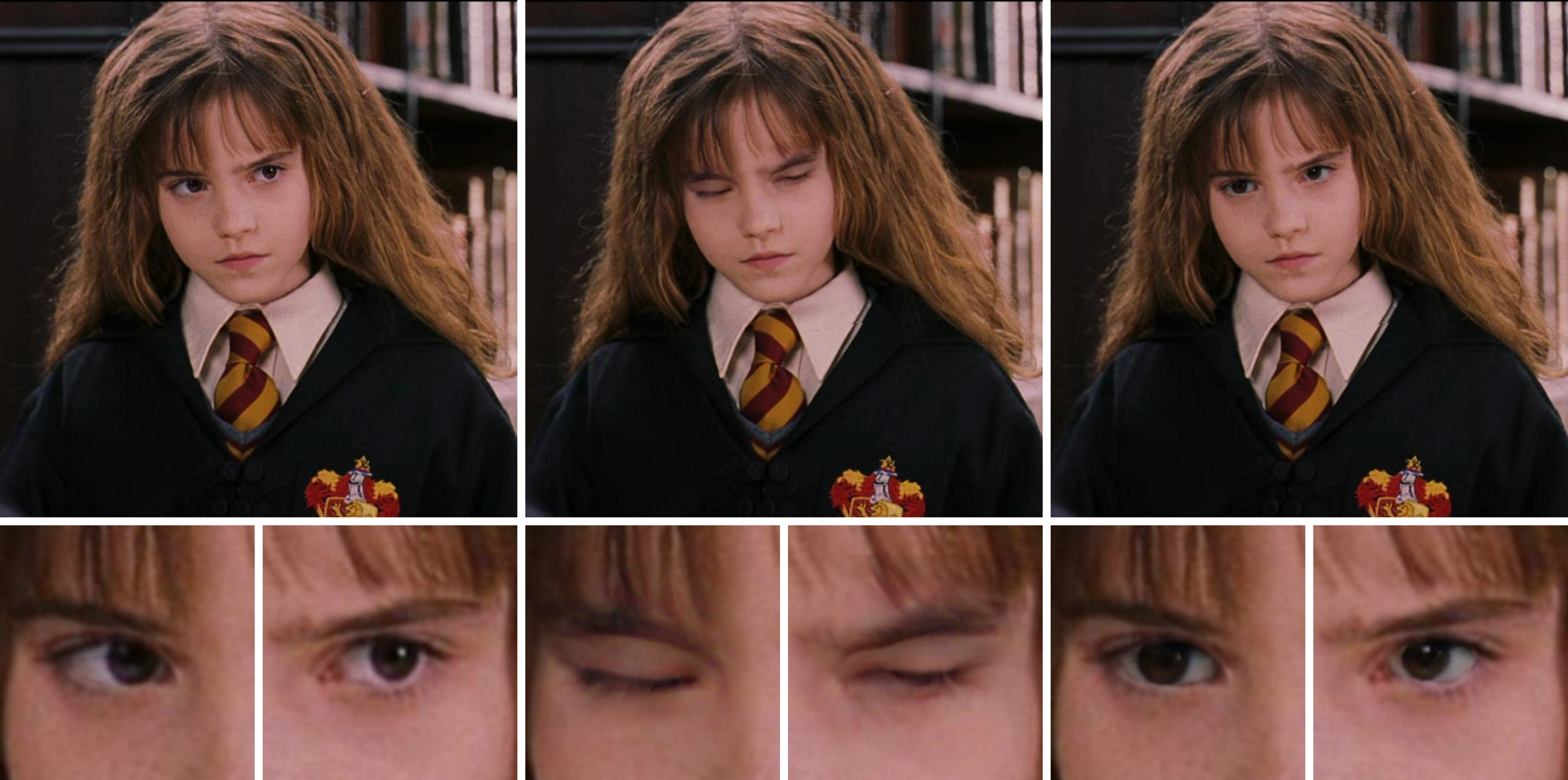}
  \onecolumn\caption{Controlling eye blink and gaze of Emma Watson as Hermione Granger in the Harry Potter series. Top Left: Source image. Top Center: Blink controlled image. Top Right: Gaze redirected image. Eye regions are magnified below each image.}\twocolumn
  \end{center}
  \end{figure}
}
\makeatother

\maketitle

\ificcvfinal\thispagestyle{empty}\fi

%%%%%%%%% ABSTRACT
\begin{abstract}
Recent developments in generative models have enabled the generation of photo-realistic human face images, and downstream tasks utilizing face generation technology have advanced accordingly. However, models for downstream tasks are yet substandard at eye control (e.g. eye blink, gaze redirection). To overcome such eye control problems, we introduce a novel framework consisting of two distinct modules: a blink control module and a gaze redirection module. We also propose a novel data augmentation method to train each module, leveraging style mixing to obtain images with desired features. We show that our framework produces eye-controlled images of high quality, and demonstrate how it can be used to improve the performance of downstream tasks.
\end{abstract}

%%%%%%%%% BODY TEXT
\section{Introduction}

\begin{figure*}
\begin{center}
\includegraphics[width=0.7\linewidth]{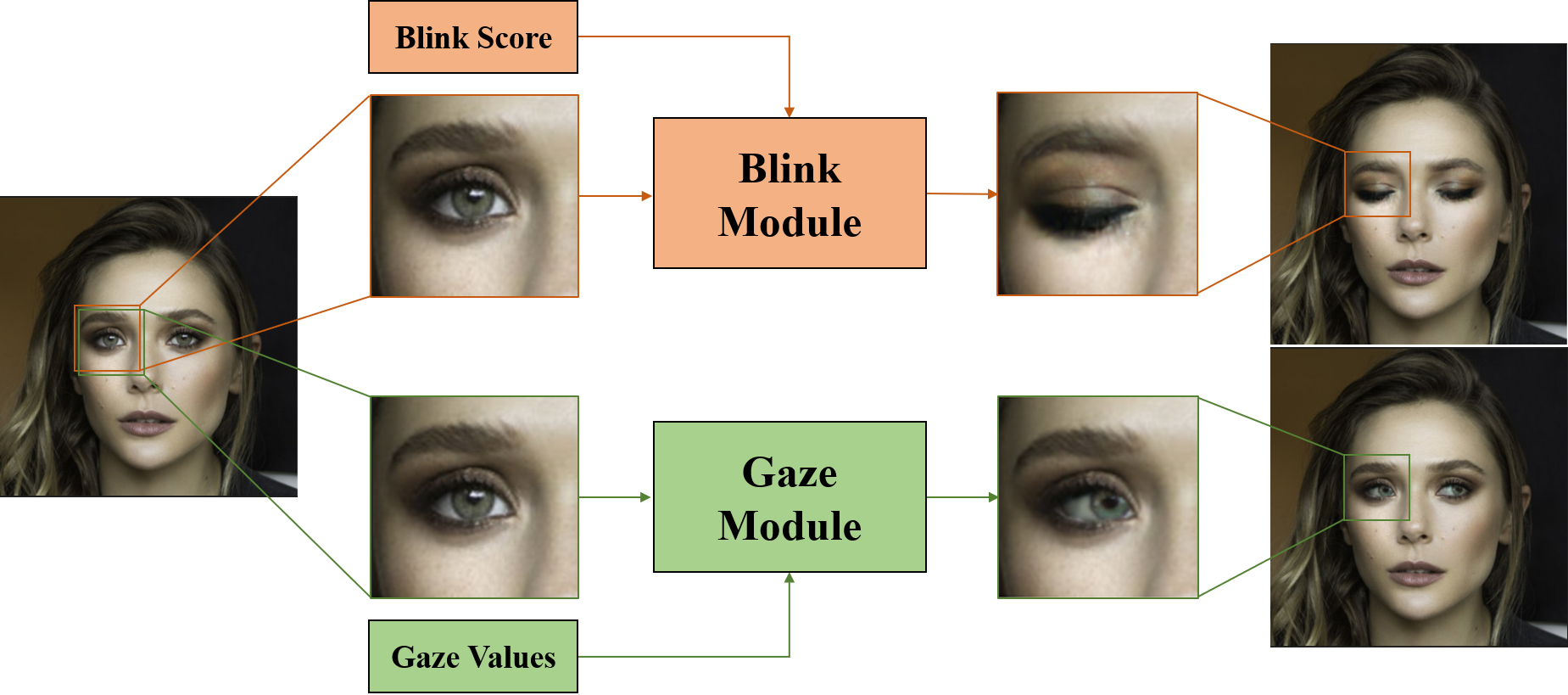}
\end{center}
\caption{Overview of the EyeBAG model}
\label{fig:short}
\end{figure*}

Interest in generative models has greatly increased over the past few years, some examples being image generation by Generative Adversarial Networks (GANs) \cite{goodfellow2020generative}, Neural Radiance Fields (NeRFs) \cite{mildenhall2021nerf, Pumarola_2021_CVPR, Barron_2021_ICCV}, and diffusion models \cite{ho2020denoising, song2020denoising, nichol2021improved}. Generative models are now more than capable of creating photorealistic images, even for complex objects such as the human face \cite{karras2019style, karras2020analyzing}. Advance in generative models has accelerated the development of downstream tasks such as face editing \cite{patashnik2021styleclip, shen2021closed}, face swapping \cite{li2019faceshifter, chen2020simswap, wang2021hififace}, and talking heads \cite{burkov2020neural, zheng2022avatar, drobyshev2022megaportraits}. However, models for face generation struggle when it comes to eye control, and such imperfection quite often creates great awkwardness and a sense of alienation. In this work, we propose a way to enhance the eye control of generative models; specifically, we enable realistic eye blinks and gaze redirection. To this end, we approach the problem from two perspectives: the dataset and the model.

Commonly used, easily accessible datasets (e.g. CelebA \cite{liu2018large}, FFHQ \cite{karras2019style}) mainly consist of images that look straight ahead directly towards the camera. In other words, ‘face images with naturally closed eyes’ and ‘face images looking in different directions’ are relatively lacking, even though such images are essential for models to learn ‘eye blink’ and ‘gaze redirection’, respectively. Furthermore, blink datasets (Eyeblink8 \cite{drutarovsky2014eye}, Researcher's Night \cite{FOGELTON201878}) and gaze datasets (RT-GENE \cite{fischer2018rt}) generally do not provide high-quality data. We feel that building a suitable dataset is greatly important for training superior blink and gaze models, hence we design a novel method to expand our dataset using style mixing. Using this method, it is possible to produce a dataset comprising pairs of open-eye and closed-eye images necessary for a model to train eye blinking. It is also possible to produce an image dataset of human faces looking at various angles, which enables high-quality gaze redirection training.

The Eye Blink And Gaze (EyeBAG) model consists of two modules: a blink control module that regulates the degree of an eye blink, and a gaze redirection module that controls the direction of gaze. Each module is trained individually with specially augmented datasets. When an input image is given, the EyeBAG model facilitates eye blinking and gaze redirection.

In summary, our main contributions are:
\begin{itemize}
\item The EyeBAG model that performs accurate blink control and gaze redirection with high performance
\item A novel method for data augmentation with style mixing
\item Data augmentation for downstream task model learning
\item A blink detection network that shows high accuracy even for videos.
\end{itemize}

%-------------------------------------------------------------------------
\begin{figure*}
\begin{center}
\includegraphics[width=.8\linewidth]{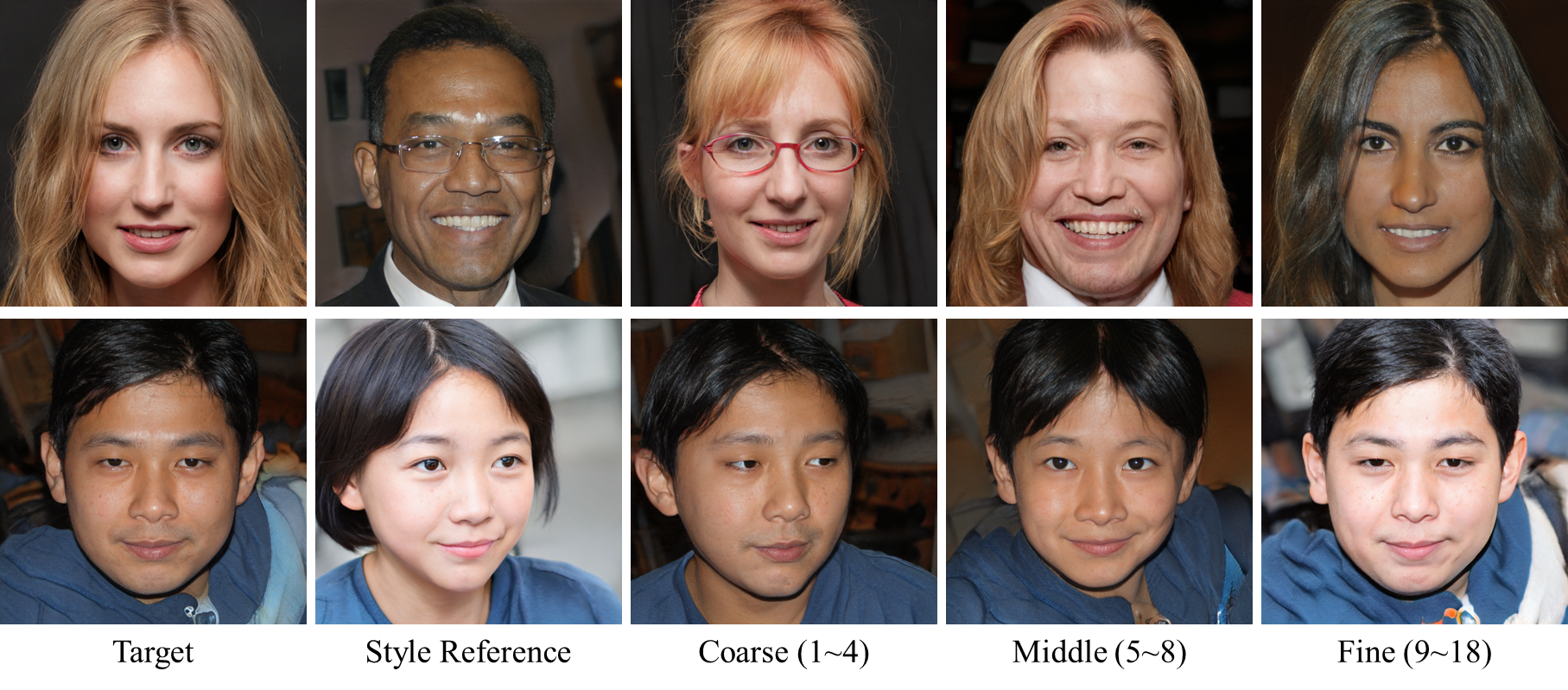}
\end{center}
\caption{Concept of style mixing. If the reference latent vector is injected in earlier layers (1~4), coarse structural features tend to mix. If the new latent vector is injected in later layers (9~18), features such as color and texture are mixed.}
\label{fig:short}
\end{figure*}

\section{Related Work}
\textbf{Eye Blink Manipulation.} Eye blink manipulation is essential for face swapping and talking head models. In these models, eye blinking is not performed well because of the challenge of capturing and synthesizing the realistic movement of eyes during the blinking process. Various tricks are used to address the issue of the eyes not blinking in these models. Deep Video Portraits \cite{kim2018deep} uses binary masks to indicate the area of the eyes and explicitly conveys the information that the eyes are closed. RAD-NeRF \cite{tang2022real} stores eye-closing frames and synthesizes them in the video sequence periodically. This model requires a video for a specific subject.

One of the key challenges in blink manipulation is to create natural-looking blinks while preserving the identity and style of the subject. InterFaceGAN \cite{shen2020interfacegan} can make eyes blink, but looks forced. Our method is able to generate natural closed-eye images through a simple U-net-based architecture.

\textbf{Eye Gaze Redirection.} Gaze redirection is a promising research area because of its potential applications such as video conferencing, virtual reality, and film-making. Various methods have been proposed for gaze redirection. Warping-based methods \cite{ganin2016deepwarp, kononenko2015learning, kononenko2017photorealistic} learn the flow field, but do not consider perceptual differences. 3D modeling-based methods \cite{banf2009example, wood2018gazedirector} use a 3D model to match texture and shape, but have many constraints for actual usage in the real world. Achieving gaze redirection through 3D transformation \cite{yang2002eye, criminisi2003gaze, zhu2011eye} proves difficult for real-world application due to the requirement of a depth map.

GAN-based gaze redirection is the most recent and realistic way of creating gaze-manipulated eyes. GAN-based methods leverage the ability of GANs that can generate realistic images with high fidelity. Gaze manipulation using GANs was first proposed by PRGAN \cite{he2019photo}. It experimentally verifies that GANs work well for non-frontal faces, but the sizes of the eye images are too small for real use. The dual in-painting model \cite{zhang2020dual} shows high quality even for wild images. It uses a novel method to create the CelebAGaze dataset. Subject Guided Eye Image Synthesis \cite{kaur2021subject} proposes a mask-based method capable of gaze redirection without gaze annotation.

\textbf{GAN-based Data Augmentation.} Attempt to apply GANs to data augmentation is increasing in various fields such as medical imaging \cite{motamed2021data}, deep fake detection \cite{rossler2019faceforensics++}, and domain adoption \cite{pinkney2020resolution}. GAN-based data augmentation generates realistic synthetic data that closely mimics real-world data, which improves the robustness and generalization of deep learning models. DatasetGAN \cite{zhang2021datasetgan} proposes an automatic procedure to generate massive datasets of high-quality semantically segmented images. It generates datasets for image segmentation tasks, which include pixel-level labelling of human face components. StyleGAN \cite{karras2019style} is able to create high-resolution images and has the ability to blend images. In this work, we use StyleGAN to create various eye conditions and use them for training the blink control module and the gaze redirection module.

%-------------------------------------------------------------------------
\section{Method}

\subsection{Style Mixing for Data Augmentation}

\textbf{Concept of Style Mixing.} Style mixing, supported by StyleGAN2 \cite{karras2020analyzing}, is a method of producing one image with the traits of two or more images using the latent vectors of the corresponding images. When a latent vector is injected into the StyleGAN generator as a style reference, all 18 layers generally receive the same vector. However, when style mixing is performed, selected layers receive as input the latent vector of a completely different image. The original image is transformed differently depending on the type and number of layers that receive the ‘new’ latent vector, as shown in Figure 3.

Until now, style mixing has existed only as a technique for mixing different images or for mixing the styles of images in different domains, and there are very few cases in which style mixing has been used as a tool for different purposes \cite{https://doi.org/10.48550/arxiv.2203.15958, DBLP:journals/corr/abs-2105-04932}. This paper introduces a novel method to use style mixing as a tool for dataset extension. The blink dataset is created by opening the eyes of closed eye face images, while the gaze dataset is created by shifting the gaze of various images. The two differ because the blink control module requires paired data to learn the process of eye blinking and the gaze redirection module benefits by acquiring images of extreme case gazes.

\begin{figure*}
\begin{center}
\includegraphics[width=0.9\linewidth]{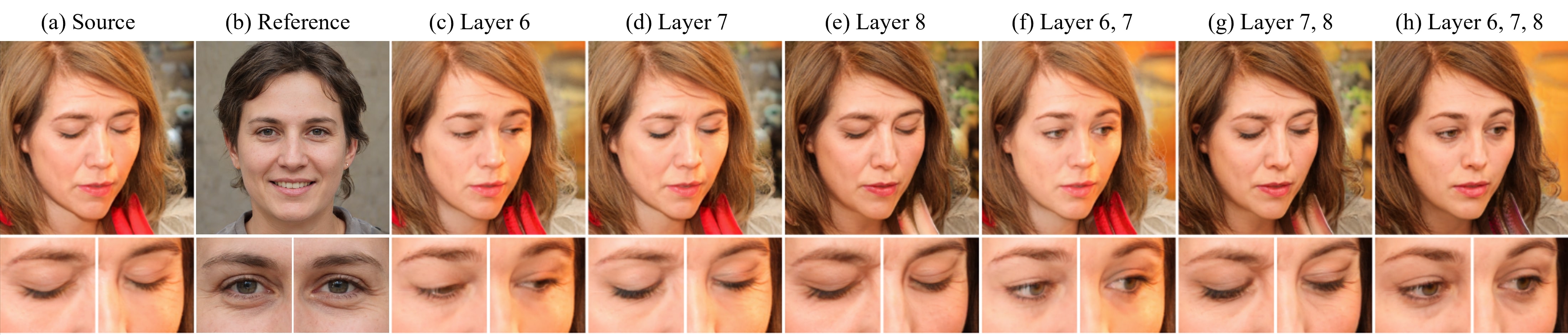}
\end{center}
\caption{Style mixing comparison (blink)}
\label{fig:short}
\end{figure*}

\begin{figure*}
\begin{center}
\includegraphics[width=0.9\linewidth]{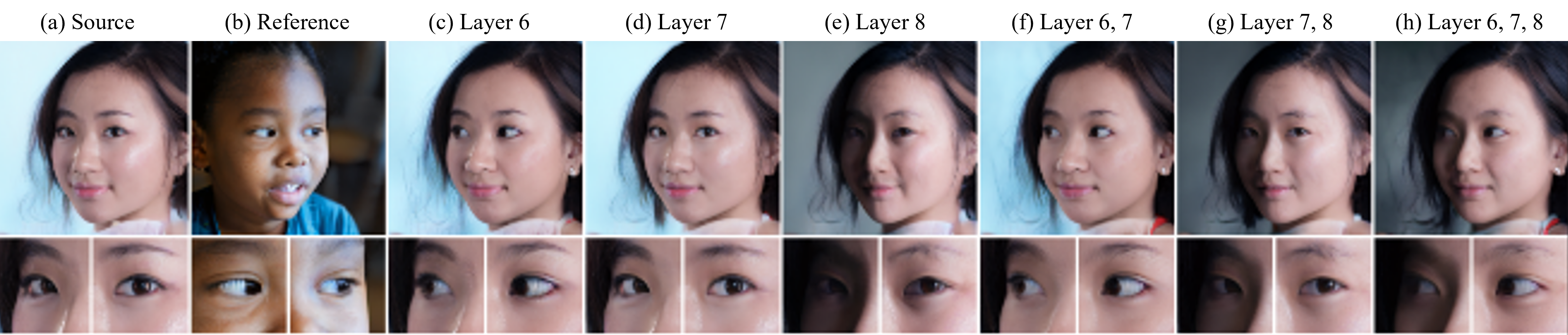}
\end{center}
\caption{Style mixing comparison (gaze)}
\label{fig:short}
\end{figure*}

\textbf{Blink Data Augmentation.} The dataset for training the blink control module consists of pairs of open-eye and closed-eye images. For the production of the dataset, closed-eye face images and uncertain-eye face images are directly selected from FFHQ random seeds (details in 4.1). The average latent vector \(w_{avg}\) is subsequently style-mixed into each image, creating an open-eye face image, which is paired with a closed-eye (or uncertain-eye) face image. Since the type and number of style-mixed layers produce different results, we conduct experiments to specify which layers should be mixed for best results. Figure 4 shows that when the average latent vector \(w_{avg}\) is style-mixed into layer 6 and 7, the result is most consistent and adequate as in Figure 4f. In Figure 4c, the eye seems to open somewhat, but it is hard to say that the eye is completely open. In Figure 4d, 4e, 4g, the eye does not seem to open at all. Style mixing all three layers as in Figure 4h gives similar results to style mixing layer 6 and 7. However, having more style-mixed layers implies more feature changes, which is undesirable for a training dataset pair. There does not seem to be a meaningful difference between Figure 4f and Figure 4h, so we decide to minimize the number of layers to be style-mixed. Thus, data production is done by mixing two layers: 6 and 7.

It is observed that people's iris colors show randomness when their eyes are open, even though there is no information about each person’s iris color and the same \(w_{avg}\) is mixed into each image. This phenomenon is due to the fact that low layers of StyleGAN express structural features, and finer features such as color or texture are expressed in higher layers \cite{karras2019style}. Since iris color is determined at layers higher than 6 and 7, various colors appear under the influence of the original latent vector. Such high iris feature diversity enables us to produce unbiased datasets for eye blink training.

\textbf{Gaze Data Augmentation.} The dataset for training the gaze redirection module is not composed of image pairs, but unpaired images that look at various angles. FFHQ seed image Figure 5b is used as a style-mix reference for gaze redirection. We perform experiments to specify which layers to style mix (Figure 5). Gaze redirection occurs when layer 6 is included for style mixing, and Figure 5c, 5f, 5h produce similar levels of gaze redirection. It is considered advantageous to change a minimal number of layers for maximal resemblance to a real human face, so only layer 6 is style mixed for dataset production. The direction of the gaze is not transferred as a result of style mixing. Figure 5b looks toward the right, while Figure 5c looks toward the left. This indicates that ‘shifted gaze’ and ‘the exact direction of the gaze’ are disentangled features, and only the former is transferred through style mixing.

\subsection{Blink Control Module}

\begin{figure*}
\begin{center}
\includegraphics[width=0.8\linewidth]{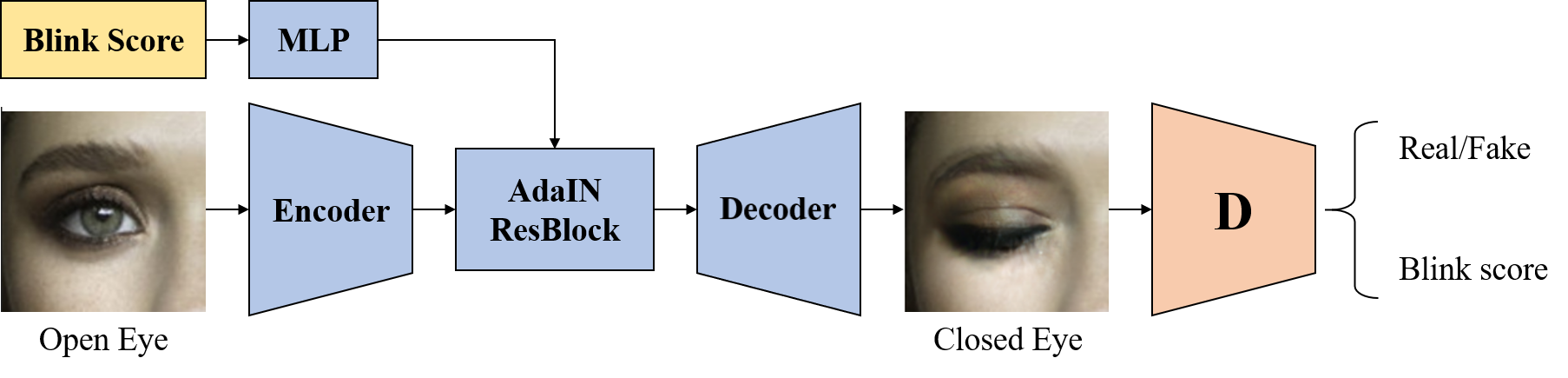}
\end{center}
\caption{Blink control module}
\label{fig:short}
\end{figure*}

\begin{figure*}
\begin{center}
\includegraphics[width=0.9\linewidth]{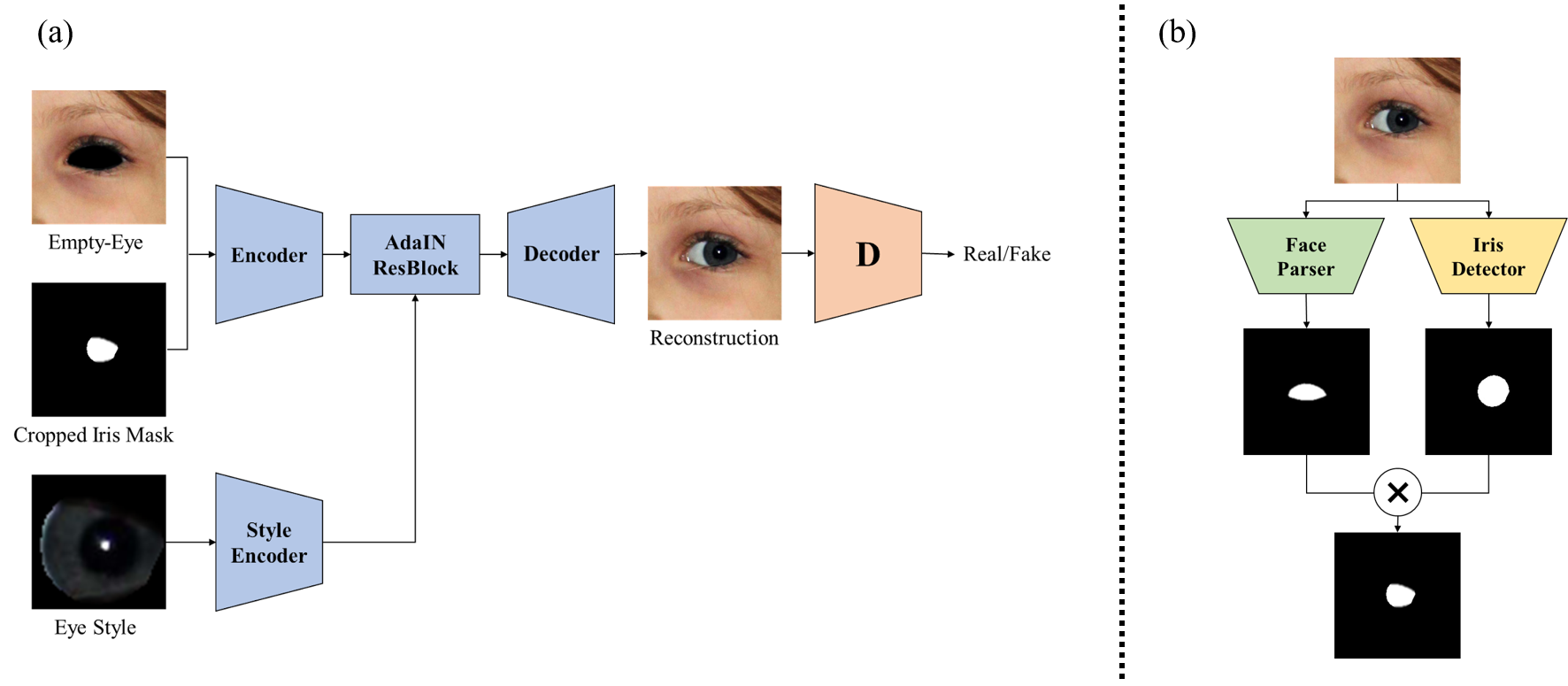}
\end{center}
\caption{Gaze redirection module}
\label{fig:short}
\end{figure*}

Let \(I_{o}\) be the open eye image and \(I_{gt}\) be the blink-controlled image. \(I_{gt}\) can be either a closed eye or an uncertain eye image. We aim to generate a blink-controlled eye image \(I_{bc}\) with \(I_{o}\) and a given blink score \(s_{b}\). 

An open eye image is encoded and input to an Adaptive Instance Normalization Residual Block (AdaIN ResBlock), along with a blink score of either 0, 0.5, or 1. Blink scores of 0 and 0.5 cause generation of closed eye images and uncertain eye images, respectively. Blink scores of 1 cause reconstruction of open eye images. These blink scores pass through a trainable multilayer perceptron (MLP). Closed eye and uncertain eye generation are learned simultaneously to effectively learn the entire blinking process, as uncertain eyes serve as the intermediate stage between being ‘open’ and ‘closed’.

\subsection{Gaze Redirection Module}

Data preprocessing is done for each eye image in the dataset to create an empty-eye image, cropped iris mask, and eye style image. The task of the module generator is to reconstruct the GT image based on the three inputs. The empty-eye image is created by masking the eye region of an image, the exact region acquired by a pre-trained face parsing network. The cropped iris mask is a mask of the visible iris region. A pre-trained face parsing network outputs the eye mask of an image, and a pre-trained iris detection network outputs the iris mask of an image. A bitwise AND operation of the two gives the cropped iris mask. The cropped iris mask and empty-eye image are concatenated and encoded before being input to an AdaIN ResBlock. The eye style image is an image of only the iris and pupil. It acts as a style reference, also input to the AdaIN ResBlock. Eye style images undergo scaling and rotation to exclude any structural information.

\begin{figure*}
\begin{center}
\includegraphics[width=0.9\linewidth]{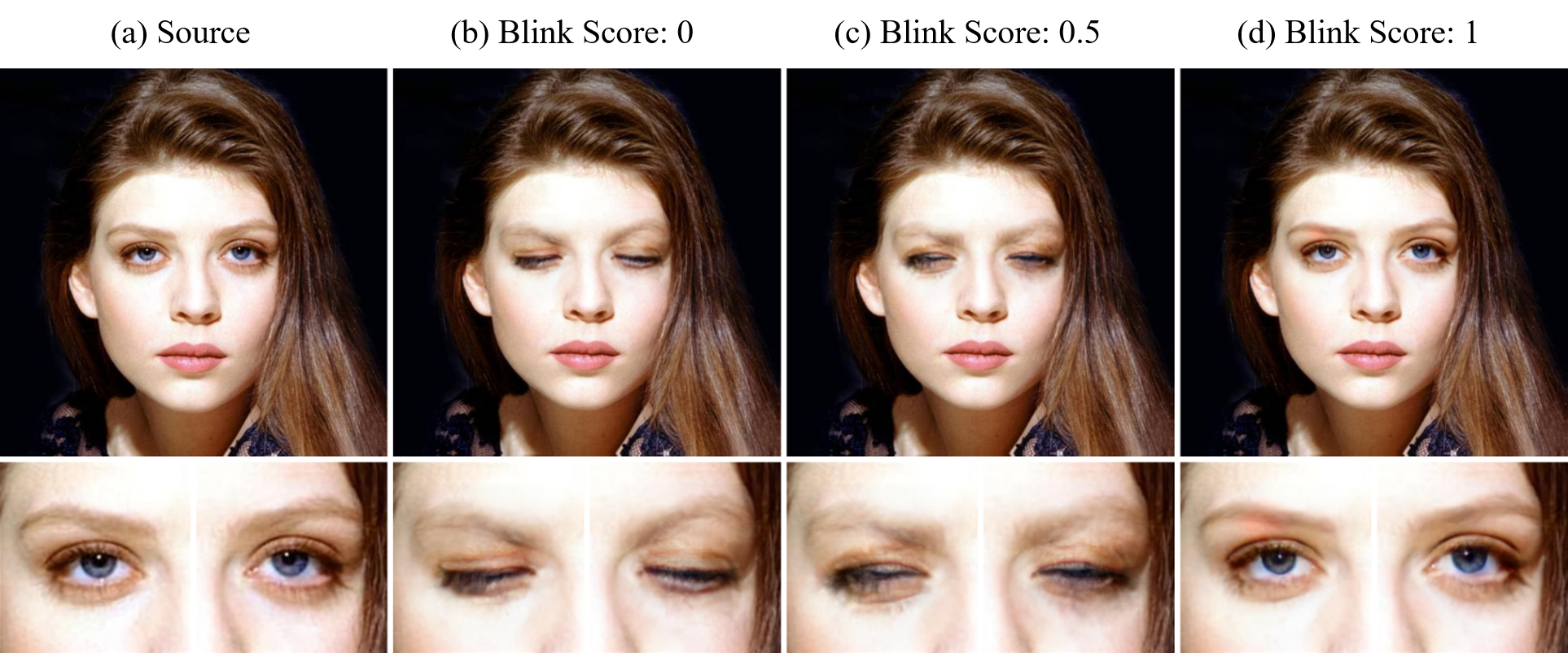}
\end{center}
\caption{Blink control results}
\label{fig:short}
\end{figure*}

\begin{figure*}
\begin{center}
\includegraphics[width=0.9\linewidth]{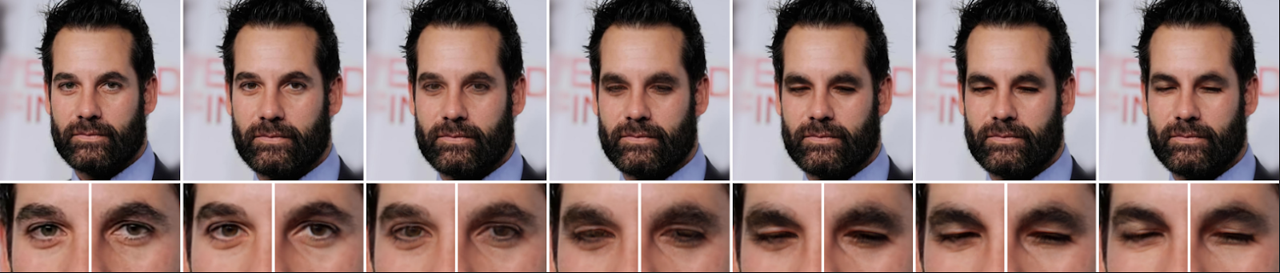}
\end{center}
\caption{Blink control results (sequential)}
\label{fig:short}
\end{figure*}

\subsection{Objectives} 
We train both the blink module and the gaze module based on supervised learning and both modules use similar loss functions. We define a reconstruction loss as a combination of pixel level L1 distance and Perceptual Image Patch Similarity (LPIPS) loss.
\[L_{recon} = \lambda_{pixel}\|I_{bc}-I_{gt}\| + \lambda_{LPIPS}\Phi_{LPIPS}(I_{bc},I_{gt})\]

The predicted blink score output from the discriminator (\(s_{o}\)) is compared with the ground truth blink score (\(s_{b}\)) input to the blink module. In addition, in order to prevent the predicted blink score from being polarized, a regularization term is added that induces the distribution of the predicted blink score to be centered at 0.5.
\[ L_{score} = \lambda_{blink}\|s_{o}-s_{b}\| + \lambda_{reg}\|s_{o}-0.5\| \]

We add the adversarial loss using a projected discriminator and a hinge loss function to produce realistic eye images similar to real data. The blink and gaze modules are finally trained with a weighted sum of the above losses as
\[ L_{blink} = L_{adv} + L_{recon} + L_{score} \]
\[ L_{gaze} = L_{adv} + L_{recon} \]

%-------------------------------------------------------------------------

\begin{figure}
\begin{center}
\includegraphics[width=0.9\linewidth]{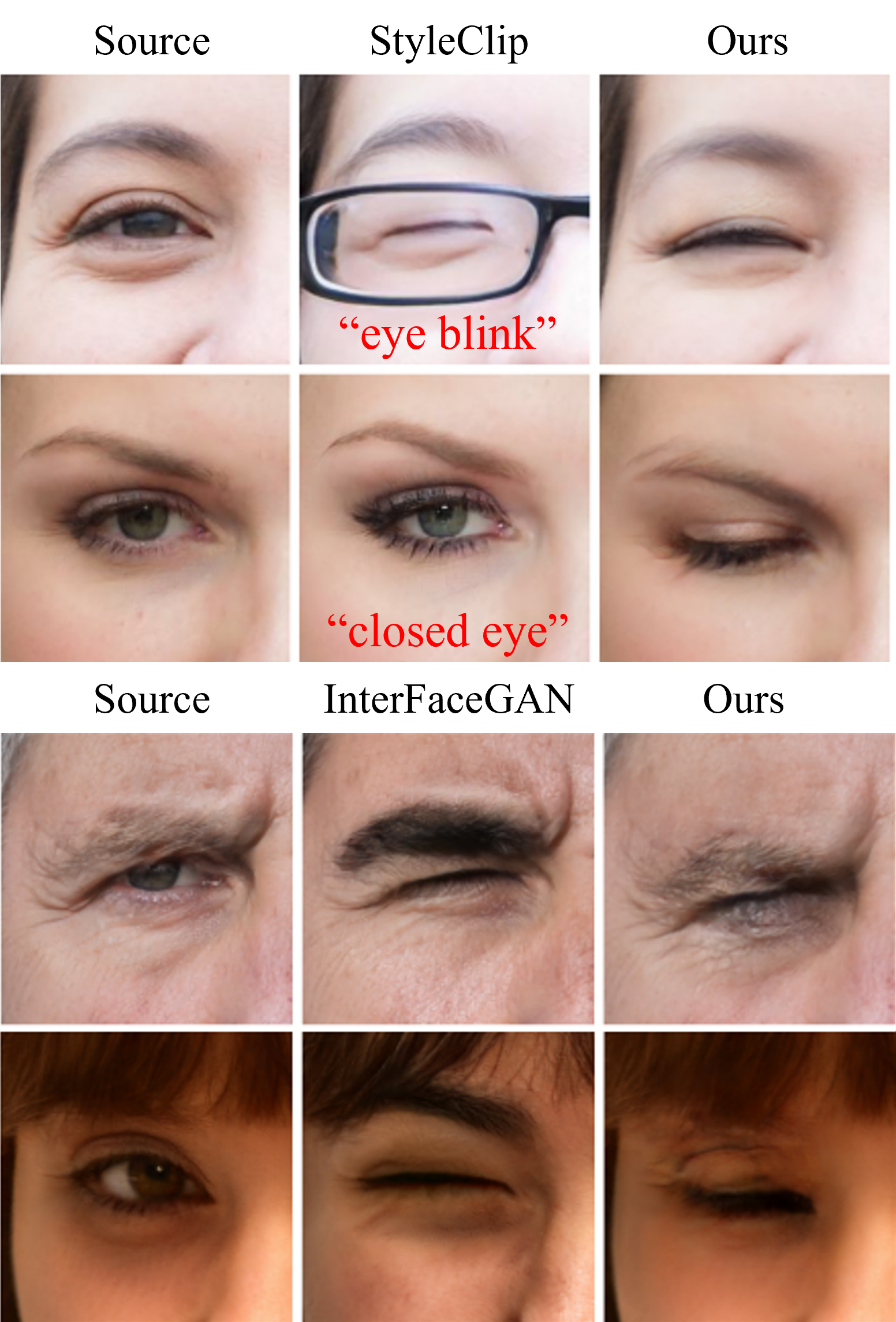}
\end{center}
\caption{Qualitative comparison}
\label{fig:short}
\end{figure}

\section{Experimental Results}

\subsection{Dataset}

The blink control module is trained by a locally generated dataset, constructed from 1120 closed eye face images and 614 uncertain eye face images selected from FFHQ random seed images. These images undergo style mixing to create open eye face images, and each image is paired with its corresponding style mixing result. All face images are then cropped to create 256*256 eye centered images, and left eye centered images are horizontally flipped for consistency with right eye centered images. Cropping is done based on the facial landmarks determined by a pre-trained facial landmark detection network.

The gaze redirection module is trained by the FFHQ dataset, along with a locally generated dataset. Style mixing is performed on 5000 FFHQ random seed images to produce a dataset, from which cases wearing eyeglasses or sunglasses are subsequently filtered out. All faces are then cropped into 256*256 eye centered images.

%-------------------------------------------------------------------------
\subsection{Implementation Details}

For both the blink control module and gaze redirection module, Adam optimizer \cite{kingma2014adam} with \(\beta_{1}=0\), \(\beta_{2}=0.999\) is used for training. Learning rates are set to 0.0001 for the generators, and set to 0.00001 for the discriminators. \(\lambda_{pixel}=10\), \(\lambda_{LPIPS}=10\), \(\lambda_{blink}=1\), \(\lambda_{reg}=0.1\) for the blink control module, and \(\lambda_{pixel}=30\), \(\lambda_{LPIPS}=30\) for the gaze redirection module.

%-------------------------------------------------------------------------
\subsection{Blink Control Module}

\begin{figure}
\begin{center}
\includegraphics[width=0.9\linewidth]{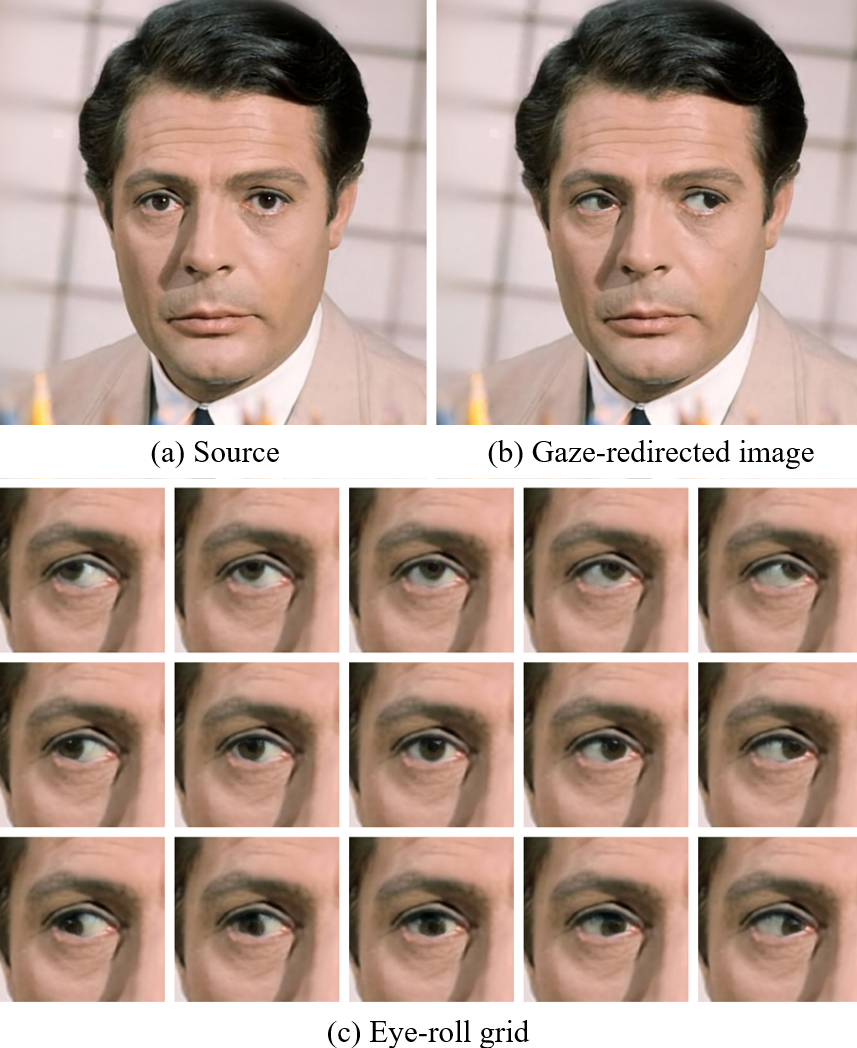}
\end{center}
\caption{Gaze redirection results}
\label{fig:short}
\end{figure}

\begin{figure}
\begin{center}
\includegraphics[width=0.9\linewidth]{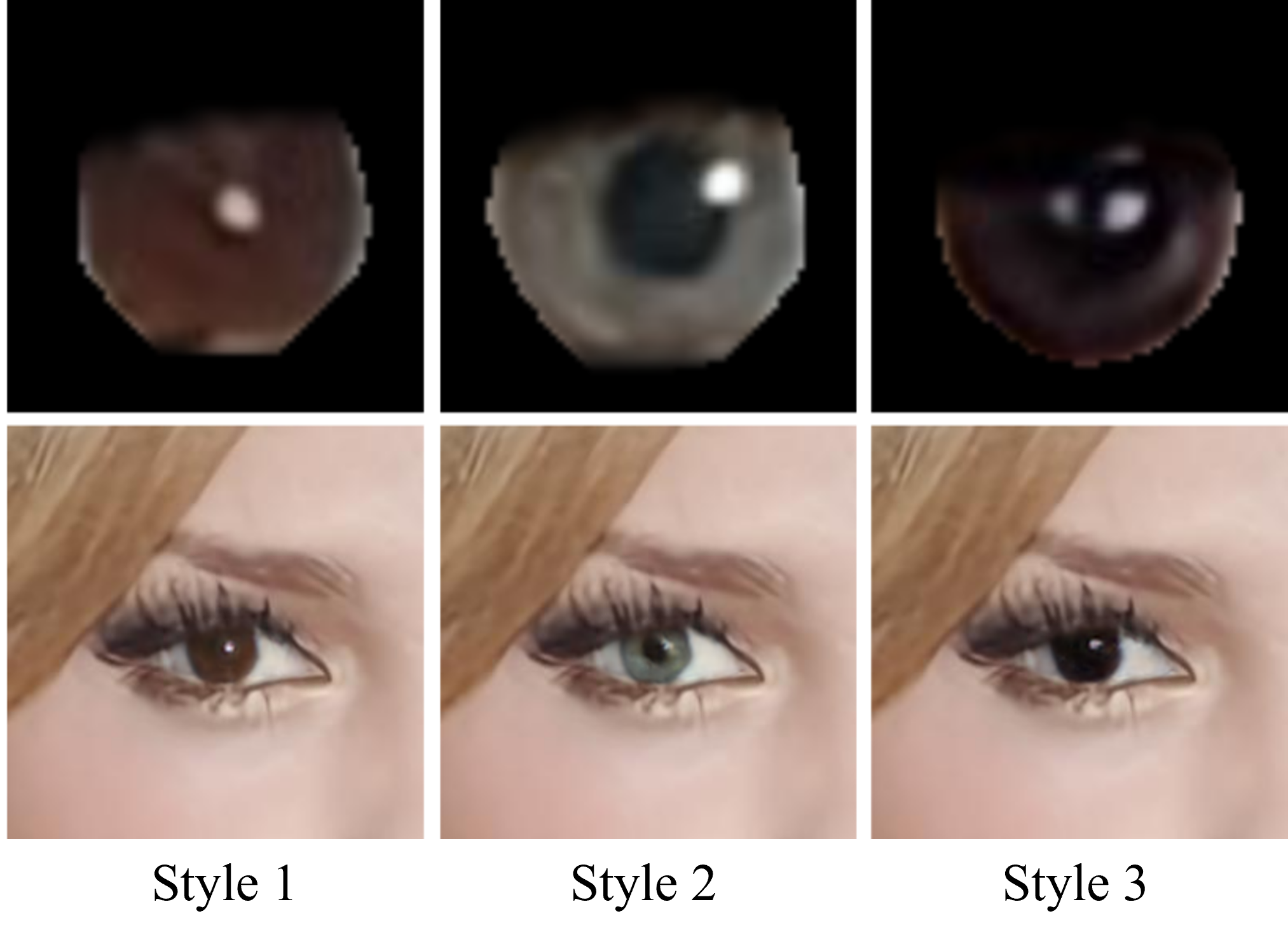}
\end{center}
\caption{Iris style transfer. For all three images on the bottom row, the same empty-eye image and cropped iris mask are used. Each image uses a different eye style image, shown on the top row}
\label{fig:long}
\end{figure}

\textbf{Qualitative Results.} Qualitative results are examined using sample images from the CelebA-HQ dataset \cite{CelebAMask-HQ}.

Transformed eye centered images output by the blink control module are attached back into the original face image using a gradient mask for natural blend. The gradient mask is created so that the weight of the output eye image is high in the eye area where large changes occur, and the weight gradually decreases in the surrounding area. This allows for minimal change in features outside the eye region.

Image generation by the blink control module is shown in Figure 8. Inputting the source image Figure 8a with a blink score of zero creates the closed eye image Figure 8b. A blink score of one creates reconstruction of the original image Figure 8d. Inputs between zero and one create uncertain eye images, as in Figure 8c. For all datasets, photo-realistic high-quality images are generated.

The blink control module is also able to express intermediate inputs other than the scores it is trained on. Figure 9 seems to depict the actual process of blinking, and the degree of the blink has a strong correlation with the input blink score. Sequential display of images with gradually decreasing blink score inputs produces a video that greatly resembles an actual blinking human.

\subsection{Gaze Redirection Module}

Input of horizontal shift value and vertical shift value is all that is needed to create a gaze-redirected image. Inputting the two values shifts the iris mask of the original image accordingly, before it is combined with the eye mask. This results in a new cropped iris mask, which in turn generates a new eye image. Any value can be input as the horizontal or vertical shift amount, allowing for gaze redirection with complete freedom. The eye centered images of Figure 11c indicate how gaze can be redirected towards any direction. Moreover, iris style is preserved with high quality, and iris styles of other face images can be transferred to the original image. Figure 12 visualizes the effects of inputting different iris styles to an original eye image.

Ablation study of how data augmentation through style mixing improves model performance is displayed in Figure 13. Data augmentation helps preserve the style and structure of the iris, especially in extreme cases. Thus, supplementing the FFHQ dataset with enough additional extreme cases allows for the gaze redirection module to produce eye images with higher quality and consistency.

\begin{figure}
\begin{center}
\includegraphics[width=0.9\linewidth]{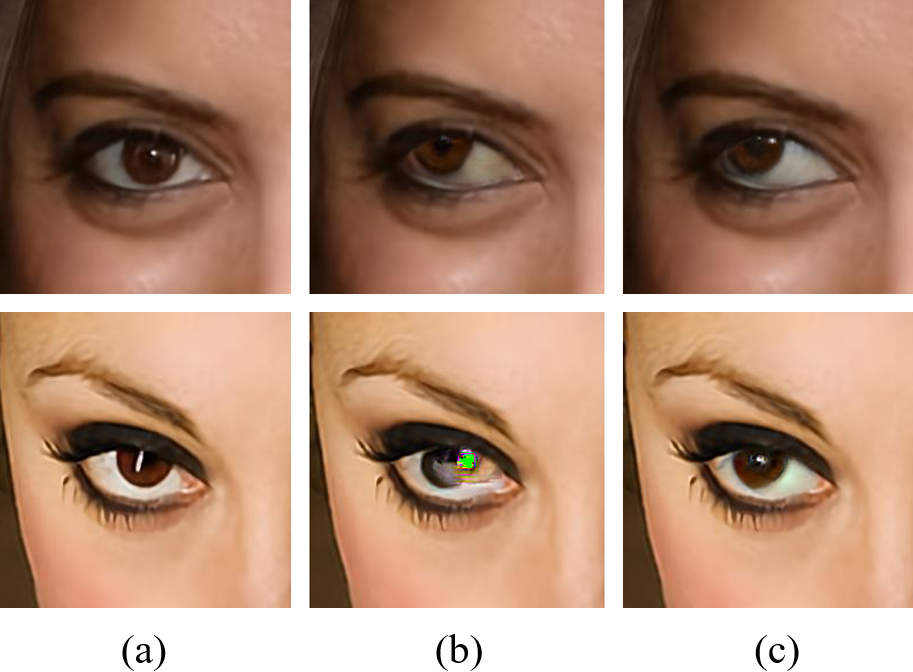}
\end{center}
\caption{Ablation study with and without style mixed data augmentation. (a) Source images from the CelebA-HQ dataset \cite{CelebAMask-HQ}. (b) Gaze redirected images without data augmentation. (c) Gaze redirected images with data augmentation. The style and structure of the iris are better preserved with data augmentation.}
\label{fig:short}
\end{figure}

%------------------------------------------------------------------------
\section{Discussion}

\textbf{Blink Detection.}
Blink score measurement with the discriminator of our blink control module is highly accurate and consistent, as described in Figure 14. Accordingly, the blink control module discriminator can be used as a pre-trained blink detection network. Figure 14 shows blink detection on frames of videos, each row portraying a series of frames from a short blink video. Blink detection is achievable at high consistency even for such videos, implying the potential usage of our blink control module.

\textbf{Performance Enhancement of Downstream Tasks.}
Images of faces with closed eyes or different gazes are relatively uncommon, causing models for downstream tasks to be substandard at eye control. Our framework produces realistic eye-controlled images of high quality, and such images can be used to improve eye control ability. Figure 15 demonstrates enhanced performance of face swapping models when supplemented by synthetic data. Models trained on a data-augmented FFHQ dataset generate better results compared to models trained without data augmentation, especially in the eye region.

\begin{figure}
\begin{center}
\includegraphics[width=0.9\linewidth]{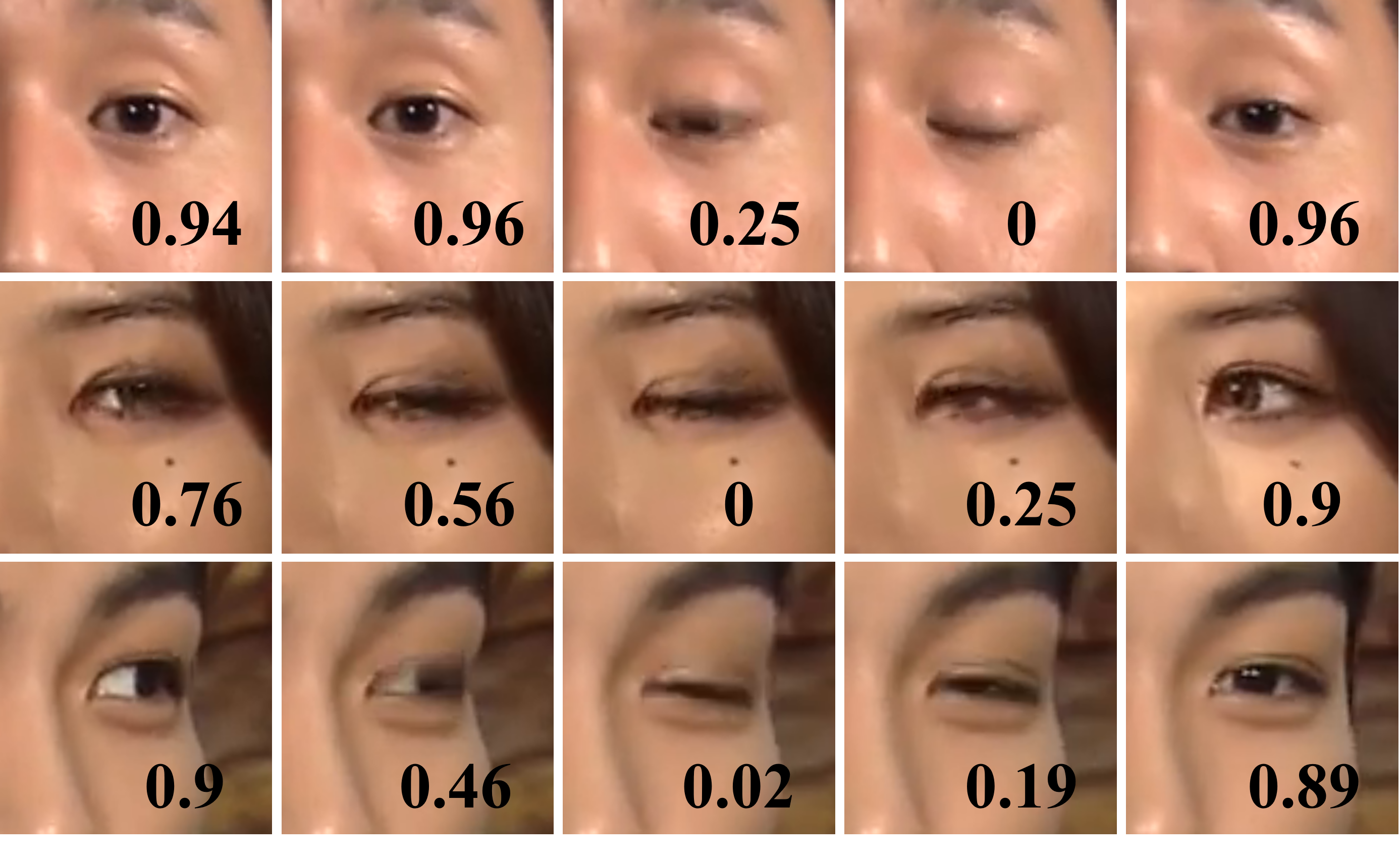}
\end{center}
\caption{Blink score measurement by the blink control module. Each row are frames extracted from the same video. Numbers indicate the output blink score, a higher score meaning the eye is more 'open' compared to a lower one. 0/1 mean closed/open eyes, respectively.}
\label{fig:short}
\end{figure}

\begin{figure}
\begin{center}
\includegraphics[width=0.9\linewidth]{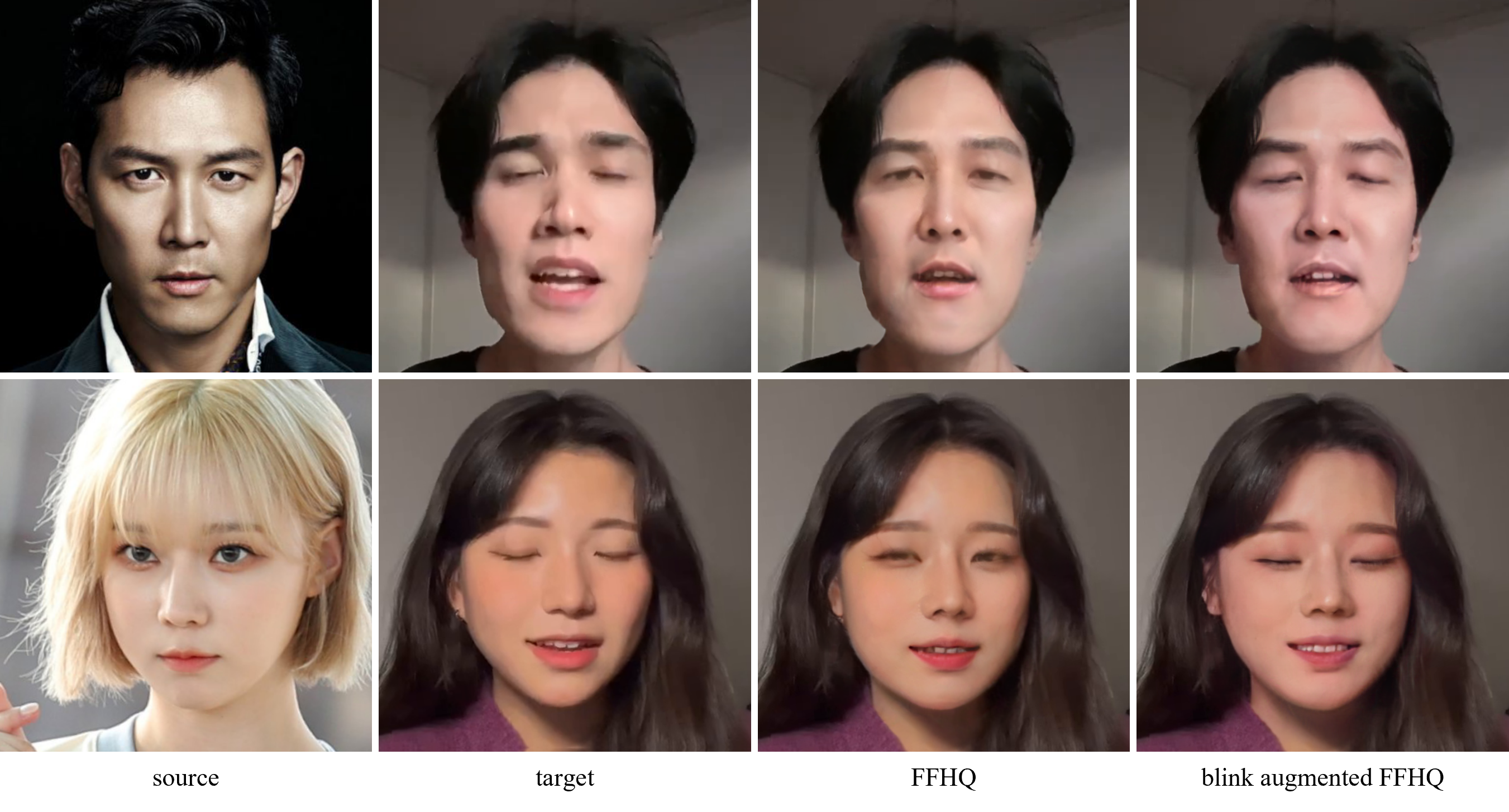}
\end{center}
\caption{Performance of face swapping downstream tasks enhanced by data augmentation. When the face swapping target has eyes closed, models trained only on FFHQ give poor results. Supplementing the model with blink image data generated by EyeBAG improves performance. Notice how the blinking state of the target image is better preserved for models trained with augmented data.}
\label{fig:short}
\end{figure}

%-------------------------------------------------------------------------
\section{Conclusion}

In this paper, we propose a data augmentation strategy and a framework for improving the eye control ability of generative models. We first introduce a novel technique for data augmentation that leverages style mixing. Locally generated datasets comprising the augmented data were used to separately train two different modules, one for blink control and one for gaze redirection. Our proposed framework, the EyeBAG model, incorporates the two modules to enable enhanced eye control capacity. The EyeBAG model generates highly realistic images of blinking faces and gaze-redirected faces, scarce data compared to the myriad of face images looking forward. We demonstrate that face swapping models struggling for realistic eye control greatly benefit from the data our model provides, and anticipate the same for similar generative models.

%------------------------------------------------------------------------

{\small
\bibliographystyle{ieee_fullname}
\bibliography{eyebag}
}

\end{document}